\documentclass[letterpaper]{article} 
\usepackage{aaai2026}  
\usepackage{times}  
\usepackage{helvet}  
\usepackage{courier}  
\usepackage[hyphens]{url}  
\usepackage{graphicx} 
\usepackage{natbib}  
\usepackage{caption} 
\usepackage{algorithm}
\usepackage{algorithmic}

\usepackage{amsmath}
\usepackage{booktabs}
\usepackage{multirow}
\usepackage{makecell}
\usepackage{amssymb}
\usepackage{graphicx}
\usepackage{subcaption}
\usepackage{amsthm}

\newtheorem{Prop}{Proposition}
\newtheorem{Lemma}{Lemma}

\usepackage{newfloat}
\usepackage{listings}
\DeclareCaptionStyle{ruled}{labelfont=normalfont,labelsep=colon,strut=off} 
\floatstyle{ruled}
\newfloat{listing}{tb}{lst}{}
\floatname{listing}{Listing}
\title{InfoCom: Kilobyte-Scale Communication-Efficient Collaborative Perception \\ with Information Bottleneck}
\author{
    Quanmin Wei\textsuperscript{\rm 1, \rm 2}, Penglin Dai\textsuperscript{\rm 1, \rm 2}\thanks{Corresponding author. }, Wei Li\textsuperscript{\rm 1, \rm 2}, Bingyi Liu\textsuperscript{\rm 3}, Xiao Wu\textsuperscript{\rm 1, \rm 2}
}
\affiliations{
    \textsuperscript{\rm 1} School of Computing and Artificial Intelligence, Southwest Jiaotong University\\

    \textsuperscript{\rm 2} Engineering Research Center of Sustainable Urban Intelligent Transportation, Ministry of Education\\
    \textsuperscript{\rm 3} School of Computer Science and Artificial Intelligence, Wuhan University of Technology\\

    wqm@my.swjtu.edu.cn, penglindai@swjtu.edu.cn, liwei@swjtu.edu.cn, byliu@whut.edu.cn, wuxiaohk@gmail.com
}

\usepackage{bibentry}

\begin{document}

\maketitle

\begin{abstract}
Precise environmental perception is critical for the reliability of autonomous driving systems. While collaborative perception mitigates the limitations of single-agent perception through information sharing, it encounters a fundamental communication-performance trade-off. Existing communication-efficient approaches typically assume MB-level data transmission per collaboration, which may fail due to practical network constraints. To address these issues, we propose InfoCom, an information-aware framework establishing the pioneering theoretical foundation for communication-efficient collaborative perception via extended Information Bottleneck principles. Departing from mainstream feature manipulation, InfoCom introduces a novel information purification paradigm that theoretically optimizes the extraction of minimal sufficient task-critical information under Information Bottleneck constraints. Its core innovations include: i) An Information-Aware Encoding condensing features into minimal messages while preserving perception-relevant information; ii) A Sparse Mask Generation identifying spatial cues with negligible communication cost; and iii) A Multi-Scale Decoding that progressively recovers perceptual information through mask-guided mechanisms rather than simple feature reconstruction. Comprehensive experiments across multiple datasets demonstrate that InfoCom achieves near-lossless perception while reducing communication overhead from megabyte to kilobyte-scale, representing 440-fold and 90-fold reductions per agent compared to Where2comm and ERMVP, respectively.
\end{abstract}

\begin{links}
    \link{Code}{https://weiquanmin.github.io/infocom}
\end{links}

\section{Introduction}

The reliability and safety of modern autonomous driving systems are significantly dependent on precise environmental perception \cite{hu2023planning, zhao2024balf, liu2024hydra, zeng2024driving, zeng2025FSDrive, ni2025recondreamer, yao2025exploring, chen2025railvoxeldet}. Collaborative perception addresses the limitations inherent to single-agent perception by complementary information exchange, thus enhancing perception performance \cite{Han2023CollaborativePI, gao2025stamp, li2024comamba, huang2024v2x}. Related strategies have been widely adopted in safety-critical tasks, including object detection \cite{xia2025learning, zhoupragmatic}, path planning \cite{qiu2022autocast}, and occupancy prediction \cite{song2024collaborative}.

\begin{figure}[t]
    \centering
    \includegraphics[width=0.99\linewidth]{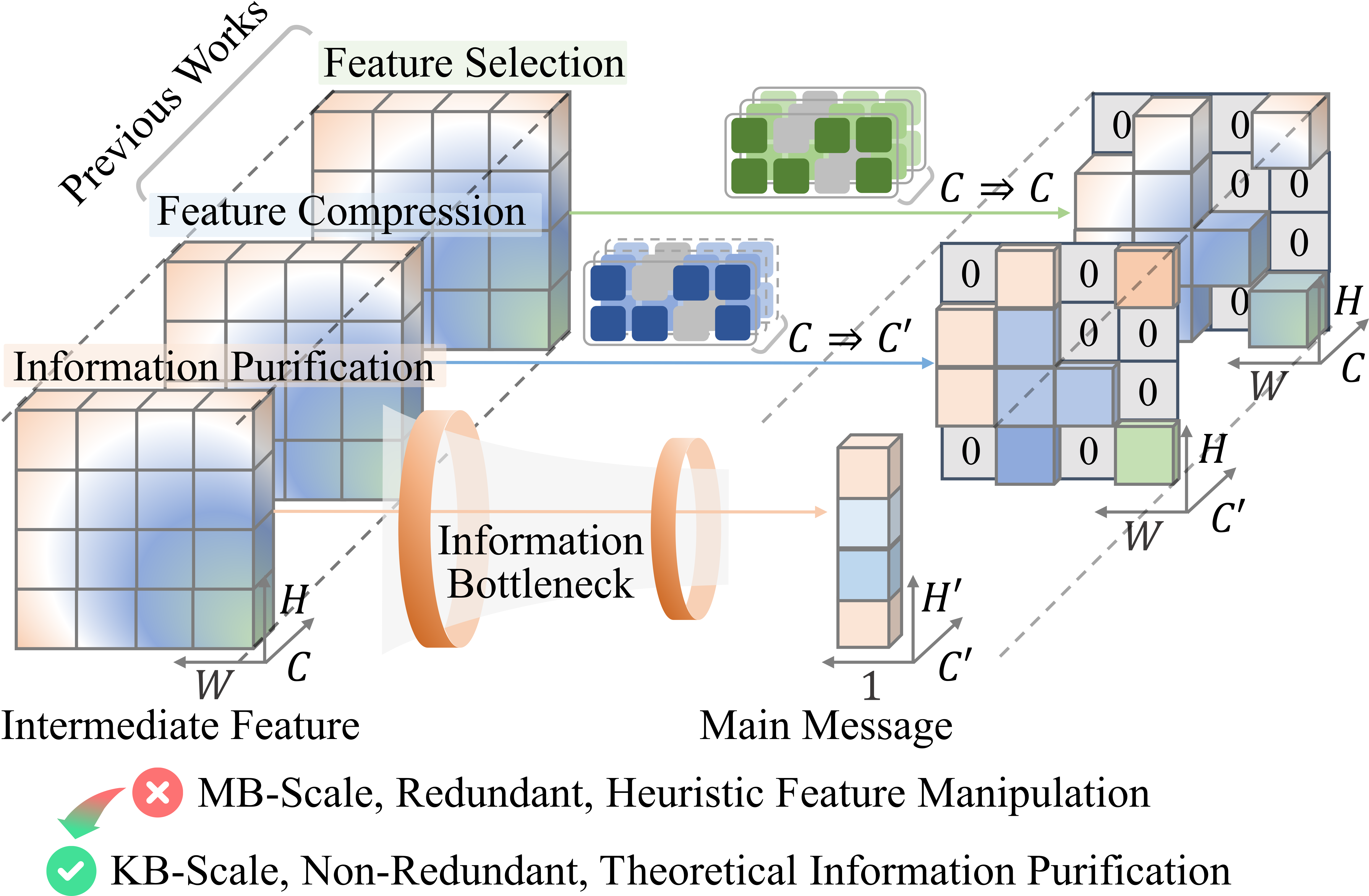}

    \caption{From redundant spatial features to essential information: InfoCom's theoretically grounded information purification ensures minimal sufficient information for perception, enabling KB-scale, near-lossless collaborative perception beyond feature manipulation.}
    \label{fig:fig1}
\end{figure}

The core trade-off of pragmatic collaboration is balancing perception performance against communication bandwidth consumption \cite{hu2024pragmatic}. Existing communication-efficient approaches primarily fall into two categories: (1) feature selection approaches, which selectively transmit critical features, but suffer from high bandwidth consumption due to their high dimensionality \cite{hu2022where2comm, zhao2023bm2cp}; and (2) feature compression approaches mapping feature into low-dimensional spaces while preserve spatial structures \cite{zhang2024ermvp, hu2024communication}. Despite progress, these solutions share significant limitations: they assume MB-level data transmission per collaboration and potentially underestimate practical network constraints. While 5G achieves $3.5$ MB/s averages in vehicular scenarios, rates may fluctuate below $0.4$ MB/s \cite{ thornton2024multi}, risking incomplete perception cycles within an acceptable time \cite{qiu2022autocast}. More fundamentally, most methods lack a theoretical analysis characterizing the communication-performance trade-off, thereby restricting their real-world applicability and optimization potential.

Here, we rethink what is a ``good" communication-efficient mechanism for addressing the above problems. In particular, Information Bottleneck (IB) provides a mathematical framework for learning a minimal sufficient representation $Z$ from observed data $X$ with target $Y$, which balances maximization of task-relevant information $I(Z;Y)$ and minimization of redundant information $I(Z;X)$ \cite{tishby2015deep}. While IB aligns intuitively with our objective, its direct application conflicts with extreme compression requirements. This incompatibility stems from the data processing inequality $ I(Z;Y) \leq I(X;Y)$ \cite{beaudry2012intuitive}, which creates an inherent tension between radical compression and high-precision perception \cite{tian2021farewell}. Consequently, this limitation reveals that existing feature-based approaches remain constrained by the high-dimensional redundancy of spatial bird’s-eye view (BEV) features, compromising task-critical information under practical network constraints.

To address this, we propose a novel information purification paradigm that directly focuses on purifying minimal sufficient information using information-theoretic criteria distinct from feature-space operations. We instantiate this paradigm as \textit{InfoCom}, an information-aware communication-efficient collaborative perception framework comprising three key modules: i) an Information-Aware Encoding module that leverages an extended IB principle to condense and preserve perception-relevant information from intermediate features, generating minimal sufficient messages; ii) a Sparse Mask Generation module that identifies spatial cues critical for collaborative decision-making through efficient filtering and quantization at negligible communication cost; and iii) a Multi-Scale Decoding module that employs mask-guided progressive reconstruction to recover perceptual information rather than simple reconstructing features.

InfoCom delivers three principal advantages: i) \textit{Extreme communication efficiency}: It maintains near-lossless perception performance while requiring only kilobyte-scale (KB) message transmission instead of the current megabyte-level (MB). ii) \textit{Plug-and-play modularity}: Its standardized design enables seamless integration with existing collaborative perception models by replacing communication layers. iii) \textit{Theoretical and empirical co-analysis}: Beyond achieving SOTA empirical performance, it establishes a pioneering theoretical analysis that supports communication-performance trade-offs, fundamentally advancing heuristic approaches.

To validate InfoCom, we conducted extensive experiments on three representative datasets: OPV2V \cite{Xu2021OPV2VAO}, V2XSet \cite{xu2022v2x}, and DAIR-V2X \cite{yu2022dair}. Experimental results show that InfoCom surpasses existing feature-based Where2comm \cite{hu2022where2comm} and ERMVP \cite{zhang2024ermvp}, with 440-fold and 90-fold reduction in communication volume. Meanwhile, InfoCom enhanced the mean AP of collaborative perception models with weaker feature extraction by 1.27\% while reducing bandwidth consumption from 34.3 MB to 2.7 KB.

\section{Related Work}

\subsection{Communication-Efficient Collaborative Perception}

Collaborative perception enables agents to share complementary information, leading to a more comprehensive understanding of the traffic environment compared to a single perception system \cite{zimmer2024tumtraf, hu2025cp, xie2025towards, tao2025directed}. This paper concentrates on the mainstream intermediate collaboration, which involves the aggregation of intermediate features \cite{xiang2024v2x, wei2025copeft}. The core challenge of this paradigm lies in balancing the communication-performance trade-off inherent in multi-agent information exchange \cite{hu2022where2comm}. Existing solutions primarily fall into two categories. \textit{Feature selection approaches} conserve bandwidth by selectively transmitting critical feature segments while discarding nonessential data \cite{liu2020when2com, liu2020who2com, wang2023core}. A representative example is Where2comm \cite{hu2022where2comm}, which employs spatial importance weighting to select key information. \textit{Feature compression approaches} employ encoding techniques to reduce high-dimensional features into compact representations while preserving spatial information for transmission \cite{hu2024communication}. For example, ERMVP \cite{zhang2024ermvp} achieves state-of-the-art communication efficiency through spatial filtering and clustering. 

\begin{figure*}[th]
    \centering
    \includegraphics[width=0.9\linewidth]{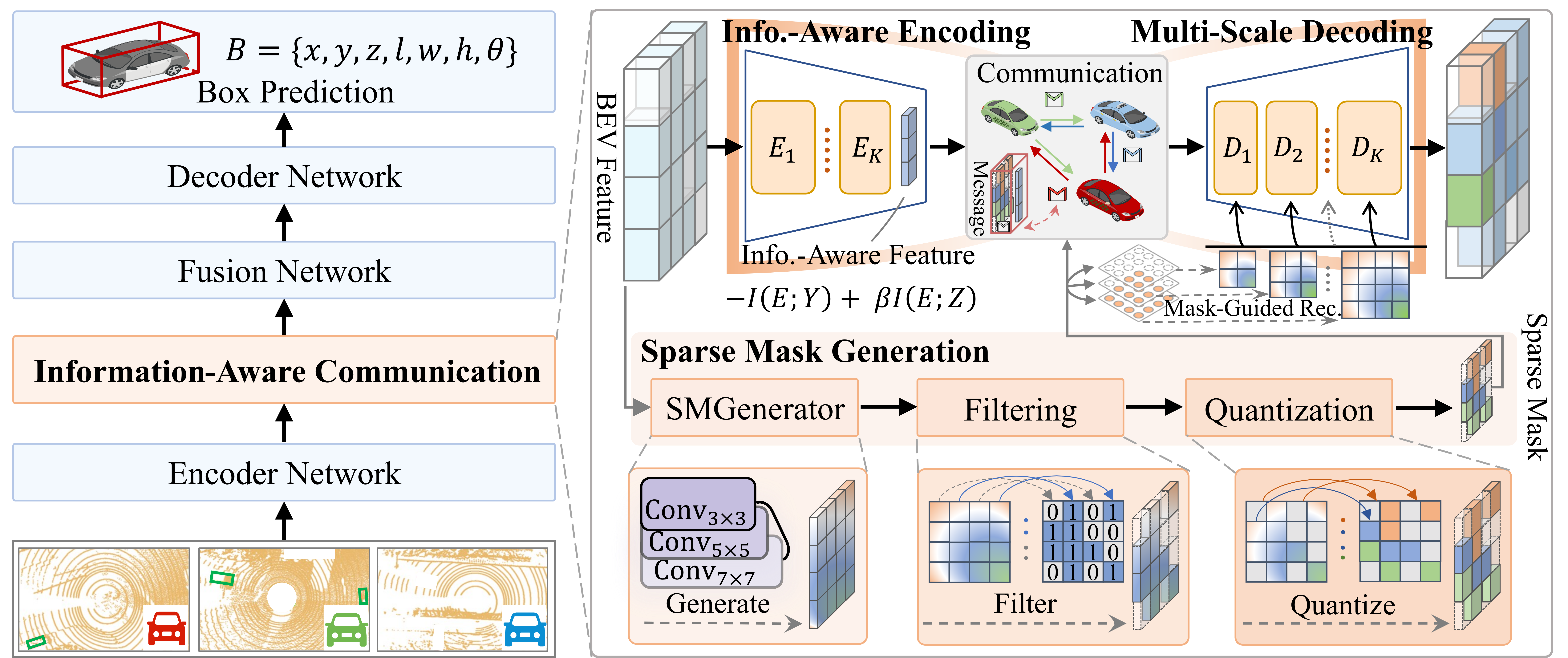}
    \caption{System overview. InfoCom is a communication-efficient collaborative perception framework based on a novel information purification paradigm, consisting of three core modules: (1) Information-Aware Encoding condenses task-critical information from high-dimensional intermediate features into minimal sufficient representations by extending the Information Bottleneck principle; (2) Sparse Mask Generation identifies essential spatial cues with minimal communication overhead; (3) Multi-Scale Decoding progressively recovers perceptual information through mask-guided reconstruction. }
    \label{fig:framework}
\end{figure*}

However, existing methods share two common limitations: i) they typically assume MB-scale bandwidth availability, whereas practical mobile networks often operate significantly below theoretical rates and render this assumption potentially unrealistic; and ii) they lack rigorous theoretical foundations for optimizing collaborative message reduction. 

\subsection{Information Bottleneck}

Our approach builds upon the Information Bottleneck (IB) theory \cite{tishby2000information, tishby2015deep}. The objective of IB is to summarize raw observation $X$ into a compact representation $Z$ while maximally preserving task-relevant information $Y$. This optimization is formally expressed as:
\begin{equation}
\label{eq:information_bottleneck}
    Z = \mathop{\mathrm{argmin}}\limits_{Z} -I(Z;Y) + \beta I(Z;X),
\end{equation}
where $I(\cdot;\cdot)$ denotes mutual information and $\beta$ serves as a Lagrange multiplier that balances information sufficiency $I(Z;Y)$ and minimality $I(Z;X)$. IB principle has been successfully applied in various domains, including representation learning and domain generalization \cite{yuan2024dynamic, wang2025out, wu2023tib}. More importantly, its core concept aligns intuitively with our goal.

However, applying the standard IB to collaborative perception encounters a fundamental limitation: its optimization objective merely trades off representational minimality against sufficiency, making it challenging to achieve extreme compression and high-accuracy perception simultaneously \cite{tian2021farewell}. To address this, we propose two key innovations: i) an extension of the IB principle using an ultra-low-dimensional feature space for preserving critical information under extreme compression, and ii) a mask-guided multi-scale decoding mechanism that ensures near-lossless perceptual accuracy through progressively spatial cues.


\section{Methodology}
\subsection{Method Overview}
\subsubsection{Problem Formulation.}

Consider a collaborative perception system comprising $N$ agents. Let $X_i$ denote the raw observation of the $i$-th agent and $Y_i$ its corresponding label. The objective of communication-efficient collaborative perception is to maximize the multi-agent perception performance under limited communication costs, that is,
\begin{equation}
\label{eq:problem_formulation}
\begin{aligned}
\mathop{\mathrm{argmax}}\limits_{\theta,\ \{\mathcal{P}_{j \to i}\}_{\substack{1 \leq i,j \leq N, j \neq i}}} \sum_{i=1}^{N} m\bigl( f_{\theta}( X_i,\{\mathcal{P}_{j \to i}\} ),\ Y_i \bigr), \\  \text{s.t.} \quad \sum^{N}_{i=1} \sum_{\substack{j=1, j \neq i}}^{N} c (\mathcal{P}_{j \to i}) \leq B,
\end{aligned}
\end{equation}
where $\mathcal{P}_{j \to i}$ denotes the learnable message transmitted from agent $j$ to $i$, $f_{\theta}$ is the collaborative perception model parameterized by $\theta$, $m$ is the perception performance metric (here focusing on 3D object detection), function $c$ quantifies bandwidth consumption, and $B$ denotes the system-level communication budget. In contrast to existing work that typically requires MB, our objective is to maximize the sufficiency of the message $\mathcal{P}_{\cdot}$ while maintaining $\frac{B}{N(N-1)}$ at the KB scale.


\subsubsection{Overall Pipeline.}
The overall workflow of InfoCom is illustrated in Fig. \ref{fig:framework}. To maintain compatibility with existing collaborative systems, our solution only requires replacing the communication layer and incorporating an Information Bottleneck regularizer into the training loss. Specifically, each agent extracts intermediate BEV features from environmental readings using a local encoder network and transforms them into a unified coordinate system with pre-shared pose. Subsequently, a customized Information-Aware Communication mechanism compresses messages at the transmitter and decompresses them at the receiver. This mechanism comprises: i) an Information-Aware Encoding that extracts minimal sufficient messages, ii) a Sparse Mask Generation that identifies spatial cues, and iii) a Multi-Scale Decoding that reconstructs processable features. Finally, the receiving agent aggregates multi-view features via a fusion network to generate 3D object detection results.

\subsection{Information-Aware Communication}
\subsubsection{Information-Aware Encoding.}
We introduce an Information-Aware Encoding (IAE) module based on the extended IB principle. To resolve IB's compression-performance dilemma, we first extend the standard Markov Chain from  $Y \rightarrow X \rightarrow Z$ to $Y \rightarrow X \rightarrow Z \rightarrow (E,M)$ where information-aware feature $E \in \mathbb{R}^{N \times D}$ satisfies $D \ll C \times H \times W$ and $Z \in \mathbb{R}^{N \times C \times H \times W}$ is intermediate feature. This extension creates a low-dimensional space for extreme compression while decoupling spatial cues as auxiliary information $M$. Accordingly, we reformulate Eq.  (\ref{eq:information_bottleneck}) to derive the new IB objective as follows:

\begin{equation}
\label{eq:information_bottleneck_new}
    E,M = \mathop{\mathrm{argmin}}\limits_{E,M} -I(E,M;Y) + \beta I(E,M;Z).
\end{equation}

In collaborative systems, the mapping from $X$ to $Z$ is handled by a fixed encoder. For the Information-Aware Encoding process $Z \rightarrow E$, we follow existing work to derive a tractable variational approximation \cite{alemi2017deep, fu2025bi}, which are instantiated via the proposed Information-Aware Encoder (IAEncoder). Specifically, for the $i$-th agent, the encoding process of IAE is represented as

\begin{equation}
 \begin{aligned}
 (\mu_{i}, \sigma_{i}) &= \text{IAEncoder}(Z_i), \\ 
 E_i &= \mu_{i} + \sigma_{i} \odot \epsilon_i, \, \epsilon_i \sim \mathcal{N}(0,I). \\ 
 \end{aligned} 
\end{equation}

The IB principle governs $E$ as minimal sufficient for perception tasks, which achieves information-aware processing. Under a $\mathcal{N}(0,I)$ prior, IAEncoder generates Gaussian parameters $\mu$ and $\sigma$ rather than complete $E$. This enables closed-form computation of the KL divergence for $I(E_i;Z_i)$ during training and straightforward implementation of the reparameterization trick for stochasticity isolation. Regarding network architecture, IAEncoder comprises three residual-like blocks designed to accommodate resource-constrained agents while ensuring robust feature extraction. Further details are provided in the Appendix.

\subsubsection{Sparse Mask Generation.}
As an essential component of InfoCom, this module compensates for critical spatial priors in perception tasks with minimal communication overhead and thereby mitigates information loss under extreme compression. According to the data processing inequality, a higher dimensional compression ratio $D / (C \times H \times W)$ implies a greater risk of task-relevant information loss. Conversely, extreme compression of $E$ frees bandwidth for transmitting auxiliary information. To this end, we introduce a Sparse Mask Generation that expands the communication unit on the sender side from solely $\mathcal{P}_i = \{E_i\}$ to $\mathcal{P}_i = \{E_i, M_i\}$. The spatial importance mask $M_i \in \mathbb{R}^{H \times W}$ is generated by the Sparse Mask Generator (SMGenerator in short) through multi-scale feature extraction, preserving perceptual cues across varying granularities:
\begin{equation}
M_i = W_{\text{proj}} (\left[ \text{Conv}_{s \times s}(Z_i) |s\in \{3,5,7\} \right]+Z_i ),
\end{equation}
where $W_{\text{proj}}$ is projection layer and $[\cdot, \cdot]$ represents channel concatenation.

We note that the initial mask $M_i$ suffers from high-entropy redundancy issue, as it is neither sparse nor compressed. Therefore, we propose a joint compression post-processing comprising two steps: filtering and quantization. In the filtering stage, our empirical observations (see Fig. \ref{fig:deeper_analysis_sub2}) indicate that only a minimal number of spatial cues benefit the task ; thus, we retain only the top-$k$ critical positions:
\begin{equation}
M_i^{\text{s}} = \text{TopK}(M_i, k), \quad k=\lfloor \alpha \cdot HW \rfloor,
\end{equation}
where $\alpha$ denotes retention ratio and is empirically set to $0.1$.

In the quantization stage, we demonstrate that spatial cues remain effective without high-precision (see Fig. \ref{fig:deeper_analysis_sub3}):
\begin{equation}
M_i^{\text{q}} = \text{Clamp}\Bigl( \text{Round}\bigl( \frac{M_i^{\text{s}}}{\delta} \bigr), 0, 2^b-1 \Bigr), \, \delta = \frac{1}{2^b-1},
\end{equation}
where $b$ is uniform quantization bit-width, defaulting to $4$.

Finally, the non-differentiability of this post-processing is resolved using the straight-through estimator like \cite{bengio2013estimating}, $\frac{\partial \mathcal{L}}{\partial M_i} \approx \frac{\partial \mathcal{L}}{\partial M_i^{\text{q}}}$. The sparse mask $M_i^{\text{q}}$ achieves efficient compression via extreme sparsity and low-precision representation. When integrated with the Multi-Scale Decoding, it delivers spatial priors at negligible communication cost while significantly mitigating information loss in $E$.

\subsubsection{Multi-Scale Decoding.}
At the receiver $k$, the efficient reconstruction of actionable features from message units $\mathcal{P}_i =\{E_i, M_i^{\text{q}}\},i \in \{1,\dots,N\}\setminus \{k\} $, is a pivotal step for translating communication efficiency into perception performance. To this end, we propose Multi-Scale Decoding (MSD), which leverages the highly compressed information-aware feature $E$ and the sparse mask $M^{\text{q}}$ to reconstruct BEV feature progressively with a focus on perceptual information. The MSD comprises three core steps. \footnote{For notational simplicity, agent ID subscripts are omitted.}

\textit{Feature Initialization.} The $E$ is expanded into a lower-resolution initial feature map ${F}^0_{\text{init}} \in \mathbb{R}^{C^0 \times H^0 \times W^0}$ via fully connected and transposed convolutional layers, where $C^0 > C$, $H^0 < H$, and $W^0 < W$. This establishes the foundation for subsequent spatial reconstruction.

\textit{Mask-Guided Modulation.} Taking ${F}^0_{\text{init}}$ as the target, the dequantization mask $M^{\text{q}} \cdot \delta \in \mathbb{R}^{H \times W}$ is downsampled to the resolution $H^0 \times W^0$ via a convolutional layer, yielding $M^0$. The features are then modulated by the mask as ${F}^0 = {F}^0_{\text{init}} \odot M^0$. This modulation directs the subsequent progressive reconstruction toward task-critical regions for enhancing perceptual information recovery.

\textit{Multi-Scale Reconstruction.} Building upon mask-guided modulation, this phase employs cascaded decoding blocks with multi-scale masks for progressive upsampling. Let $F^i$ denote the output of the $i$-th block, with resolution $C^i \times H^i \times W^i$ that satisfies $2C^i = C^{i-1}$, $H^i = 2H^{i-1}$, and $W^i = 2W^{i-1}$. After $K$ iterations, the feature map $F^K$ reaches the target resolution $C \times H \times W$ of intermediate feature $Z$.

Multi-Scale Decoding processes all messages to actionable BEV features $F_{1:N}=\left\{ F^K_i \mid i \in \{1, \dots, N\} \setminus \{k\} \right\} $ at the $k$-th agent. Finally, the resulting set $F_{1:N} \cup \{ Z_k \}$ is fed directly into a standard fusion network to aggregate multi-source information, ensuring compatibility with collaborative perception systems.

\subsection{Overall Loss}

To enable end-to-end training, we follow existing works by decomposing the IB objective in Eq.  (\ref{eq:information_bottleneck_new}) into a supervised loss and a regularization term \cite{wu2023tib, chen2023variational}. The overall loss function is shown as follows:

\begin{equation}
\label{equa:loss}
\mathcal{L} = \mathcal{L}_{\text{detect}} + \beta  \text{KL}\big(p(E|Z) \big\| r(E)\big),
\end{equation}
where $\mathcal{L}_{\text{detect}}$ denotes the standard detection loss consistent with existing collaborative perception models \cite{Lu2022RobustC3}, the $\text{KL}(\cdot)$ divergence term instantiates the IB regularization with $p(E|Z)$ representing the stochastic mapping $Z \to E$ parameterized by the IAEncoder. Under Gaussian prior assumption $r(E) = \mathcal{N}(0,I)$ for variable $E$, this regularization term admits a closed-form solution.

\begin{table*}[t]
    \setlength{\tabcolsep}{3mm}
    \centering
    \begin{tabular}{c|c|c|c|c|c}
    \toprule
        Dataset & Comm. Method & Comm. Volume & AP@30 & AP@50 & AP@70 \\ 
    \midrule
        \multirow{6}{*}{\makecell[c]{OPV2V\\ \cite{Xu2021OPV2VAO}}} & No Collaboration & 0 & 0.8665 & 0.8464 & 0.7288 \\ 
        ~ & Late Collaboration & 6.250 KB & 0.9622 & 0.9499 & 0.8495 \\ 
        ~ & Standard Colla. \cite{Lu2022RobustC3} & 34.375 MB & 0.9709 & 0.9653 & 0.9229 \\ 

        ~ & Where2comm \cite{hu2022where2comm} & 3.439 MB & 0.9548 & 0.9463 & 0.8820 \\ 
        ~ & ERMVP \cite{zhang2024ermvp} & \underline{0.741 MB} & \underline{0.9618} & \underline{0.9557} & \underline{0.9127} \\ 
        ~ & InfoCom (Ours) & \textbf{7.875 KB} & \textbf{0.9702} & \textbf{0.9650} & \textbf{0.9202} \\
    \midrule
        \multirow{6}{*}{\makecell[c]{V2XSet \\ \cite{xu2022v2x}} } & No Collaboration & 0 & 0.8032	& 0.7719 & 0.6222 \\ 
        ~ & Late Collaboration & 6.250 KB & 0.9321 & 0.9076 & 0.7120 \\ 
        ~ & Standard Colla. \cite{Lu2022RobustC3} & 34.375 MB & 0.9317 & 0.9212 & 0.8426 \\ 
        
        ~ & Where2comm \cite{hu2022where2comm} & \underline{3.439 MB} & \underline{0.8834} & \underline{0.8604} & \underline{0.7417} \\ 
        ~ & ERMVP \cite{zhang2024ermvp} & / & OOM & OOM & OOM \\ 
        ~ & InfoCom (Ours) & \textbf{7.875 KB} & \textbf{0.9360} & \textbf{0.9273} & \textbf{0.8488} \\ 
    \midrule
        \multirow{6}{*}{\makecell[c]{DAIR-V2X \\ \cite{yu2022dair}}} & No Collaboration  & 0 & 0.7278	& 0.6844 & 0.5665 \\ 
        ~ & Late Collaboration & 6.250 KB & 0.7993 & 0.6709 & 0.4708 \\
        ~ & Standard Colla. \cite{Lu2022RobustC3} & 24.609 MB & 0.8294 & 0.7843 & 0.6353 \\ 
         
        ~ & Where2comm \cite{hu2022where2comm} & 2.462 MB & 0.8048 & 0.7539 & 0.6070 \\ 
        ~ & ERMVP \cite{zhang2024ermvp} & \underline{0.531 MB} & \underline{0.8217} & \textbf{0.7791} & \underline{0.6324} \\ 
        ~ & InfoCom (Ours) & \textbf{5.922 KB} & \textbf{0.8228} & \underline{0.7789} & \textbf{0.6385} \\ 
    \bottomrule
    \end{tabular}
    \caption{Performance comparison on three representative collaborative perception datasets. All communication-efficient methods are built upon CoAlign, with the best and second-best results highlighted in bold and underlined, respectively.}
    \label{table:main_result_1}
\end{table*}

\section{Theoretical Analysis}

Here, we provide a theoretical analysis, where Lem. \ref{lemma:lemma1_extended} establishes the noise bound for the compression process, and Prop. \ref{prop:prop1_extended} demonstrates that InfoCom achieves communication efficiency, information retention, and noise suppression.

\begin{Lemma}
\label{lemma:lemma1_extended}
\textbf{(Noise Suppression Bound of Collaborative Features)}
Consider an agent's observation $X$, intermediate feature $Z$, message unit $\mathcal{P}=\{E,M\}$ consisting of the information-aware feature $E$ and the sparse mask $M$, and the perceptual target $Y$ with task-irrelevant noise $Y_N$ satisfying $Y\perp Y_N$. 
Under the Markov chain $(Y,Y_N)\!\rightarrow\!X\!\rightarrow\!Z\!\rightarrow\!(E,M)$, the following inequality holds:
\begin{equation}
I(E,M;Y_N)\;\le\;I(E,M;Z)\;-\;I(E,M;Y).
\end{equation}
Furthermore, due to the entropy constraint of $M$,
\begin{equation}
\scriptsize
I(E,M;Z)\;\le\;I(E;Z)+H(M)\;\le\;I(E;Z)+\log\binom{HW}{k}+k\,b,
\label{equa:bound}
\end{equation}
where $k=\lfloor \alpha \cdot HW \rfloor$ is the number of retained spatial positions and $b$ is the quantization bit width.
\end{Lemma}

\begin{proof}
Since $(E,M)$ is a function of $Z$, the data processing inequality applied to the Markov chain $(Y,Y_N) \rightarrow Z \rightarrow (E,M)$ gives $I(E,M;Z)\!\ge\!I(E,M;Y,Y_N)$.
Decomposing the mutual information yields
$I(E,M;Y,Y_N)=I(E,M;Y_N)+I(E,M;Y|Y_N)$.
Since $Y\!\perp\!Y_N$, we have $H(Y|Y_N)=H(Y)$, and thus
$I(E,M;Y|Y_N)\!\ge\!I(E,M;Y)$. Substituting and rearranging gives
$I(E,M;Y_N)\!\le\!I(E,M;Z)-I(E,M;Y)$.
Finally, decomposing $I(E,M;Z)=I(E;Z)+I(M;Z|E)$
and applying $I(M;Z|E)\!\le\!H(M)\!\le\!\log\binom{HW}{k}+k b$
establishes the stated bound.
\end{proof}

\begin{Prop}
\label{prop:prop1_extended}
\textbf{(Theoretical Foundations of Information-Aware Communication in Collaborative Perception)} The InfoCom achieves communication-efficient collaborative perception through three interconnected mechanisms with theoretical foundations: (1) bandwidth reduction via filtering and quantization, (2) preservation of task-relevant information, and (3) suppression of task-irrelevant noise.


\end{Prop}

\begin{proof}
(1) \textit{Bandwidth reduction:} The $\mathcal{P}=\{E,M\}$ achieves substantial bandwidth reduction through explicit design choices: $E\in\mathbb{R}^D$ with $D\ll C\times H\times W$ enables dimension reduction, while the sparse mask $M$ selects only $k\ll H \times W$ positions with $b$-bit quantization. Lem.~\ref{lemma:lemma1_extended} provides the information-theoretic foundation through the bound (\ref{equa:bound}), which constrains the fundamental entropy of $\mathcal{P}$.

(2) \textit{Perceptual preservation:} The detection loss $\mathcal{L}_{\text{detect}}$ implicitly maximizes $I(E,M;Y)$, ensuring preservation of task-relevant information for perceptual accuracy.

(3) \textit{Noise suppression:} From Lem.~\ref{lemma:lemma1_extended}, $I(E,M;Y_N)\le I(E,M;Z)\!-\!I(E,M;Y)$; thus, minimizing $I(E,M;Z)$ via IB regularization and maximizing $I(E,M;Y)$ via detection loss jointly suppress noise. Furthermore, quantization processes provide additional regularization through the data processing inequality, enhancing robustness.
\end{proof}


\begin{table*}[t]
    \setlength{\tabcolsep}{3mm}
    \centering
    \begin{tabular}{c|c|ccc|ccc}
    \toprule
        \multicolumn{2}{c}{} & \multicolumn{3}{c}{AttFuse \cite{Xu2021OPV2VAO}} & \multicolumn{3}{c}{MKD-Cooper \cite{li2023mkd}} \\
    \midrule
        Comm. Method & Comm. Volume & AP@30 & AP@50 & AP@70 & AP@30 & AP@50 & AP@70 \\ 
    \midrule
        Standard Colla. & 34.375 MB & 0.9546 & 0.9327 & 0.8039 & 0.9575 & 0.9360 & 0.8153 \\ 
        Where2comm & 3.438 MB & 0.9173 & 0.9004 & \underline{0.8023} & \underline{0.9134} & \underline{0.8974} & \underline{0.7957} \\ 
        ERMVP & \underline{0.701 MB} & \underline{0.9212} & \underline{0.9041} & 0.7917 & 0.9112 & 0.8946 & 0.7848 \\ 
        InfoCom (Ours) & \textbf{2.718 KB} & \textbf{0.9575} & \textbf{0.9360} & \textbf{0.8153} & \textbf{0.9640} & \textbf{0.9490} & \textbf{0.8340} \\ 
    \bottomrule
    \end{tabular}
    \caption{Evaluation of communication-efficient methods using alternative collaborative perception models on OPV2V dataset.}
    \label{table:main_result_2}
\end{table*}

\section{Experiments}
\subsection{Experimental Settings}
\subsubsection{Datasets and Evaluation Metrics.}
We follow existing work \cite{wei2025copeft} and conduct experiments on three representative collaborative perception datasets: the simulated OPV2V \cite{Xu2021OPV2VAO} and V2XSet \cite{xu2022v2x}, and the real-world DAIR-V2X \cite{yu2022dair}. For DAIR-V2X, we follow standard practices to extend annotation coverage \cite{Lu2022RobustC3}. 3D object detection performance is evaluated using Average Precision (AP) at Intersection over Union (IoU) thresholds of 0.3, 0.5, and 0.7. Communication volume is measured in human-readable units to estimate actual transmission requirements per agent.

\begin{figure}[t]
    \centering
    \includegraphics[width=0.99\linewidth]{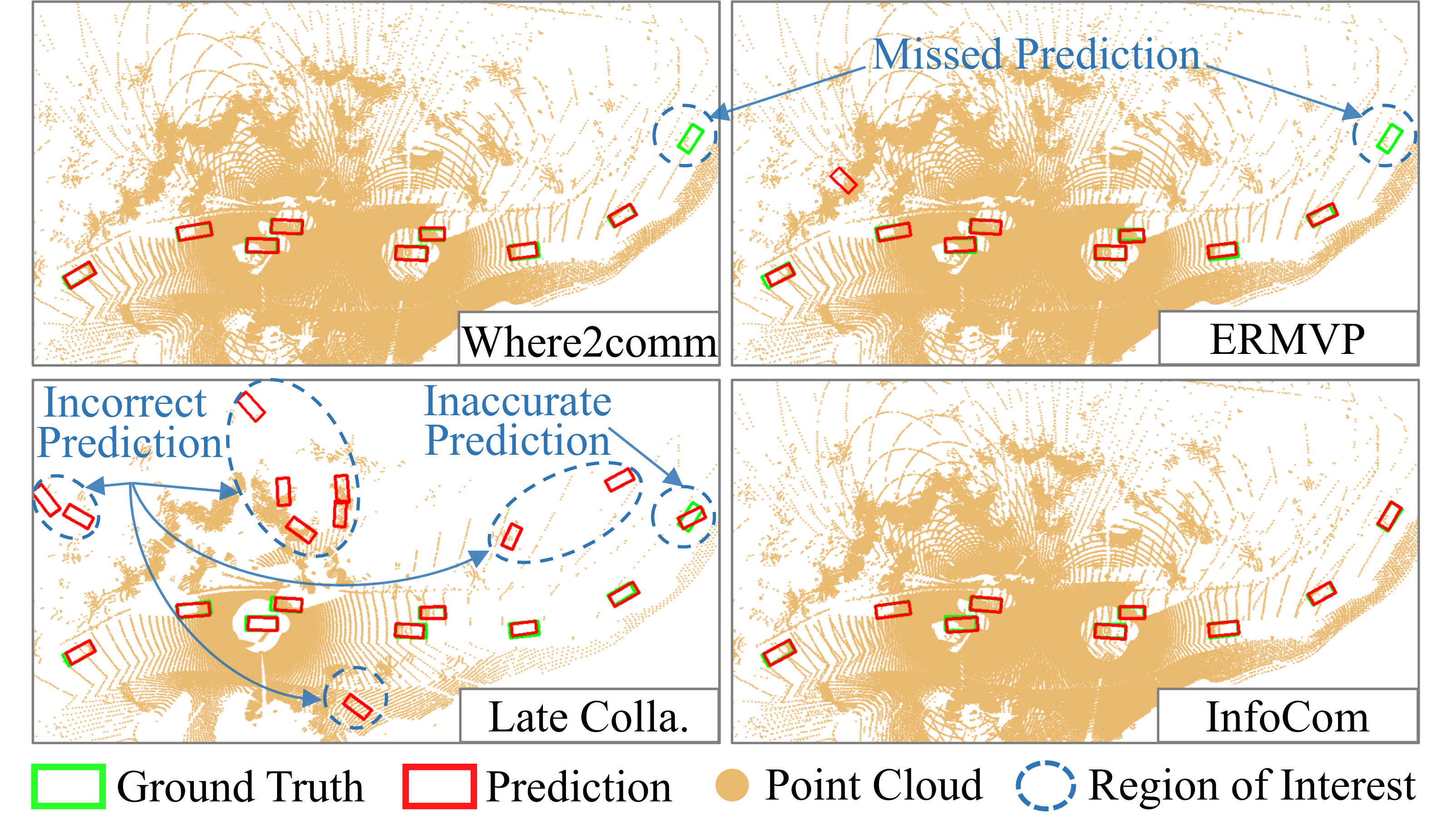}
    \caption{Qualitative comparison on OPV2V dataset.}
    \label{figure:qualitative_analysis}
\end{figure}

\subsubsection{Baselines.}
InfoCom's communication efficiency is validated against SOTA feature-based alternatives, Where2comm \cite{hu2022where2comm} and ERMVP \cite{zhang2024ermvp}. For fairness, all non-communication components and customizable settings of the base collaborative perception model remain fixed. The default base employs the multi-scale CoAlign framework \cite{Lu2022RobustC3}, while single-scale models, AttFuse \cite{Xu2021OPV2VAO} and MKD-Cooper \cite{li2023mkd}, validate cross-model applicability. We concurrently report performance for standard intermediate, late, and no collaboration as baseline references. More experiment details are provided in the Appendix.

\begin{figure*}[t]
    \centering
    \begin{subfigure}[b]{0.3\textwidth}
        \centering
        \includegraphics[width=\textwidth]{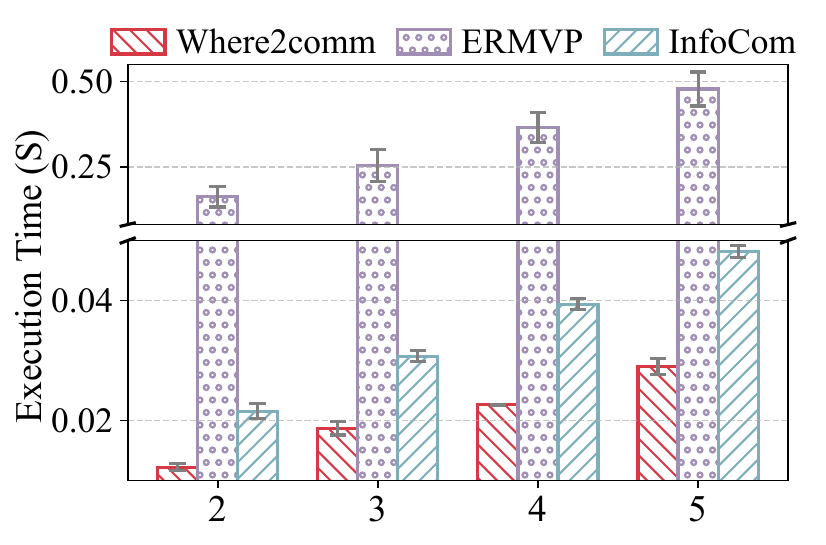}
        \caption{Execution time w.r.t agent density}
        \label{fig:deeper_analysis_sub1}
    \end{subfigure}
    \hfill
    \begin{subfigure}[b]{0.33\textwidth}
        \centering
        \includegraphics[width=\textwidth]{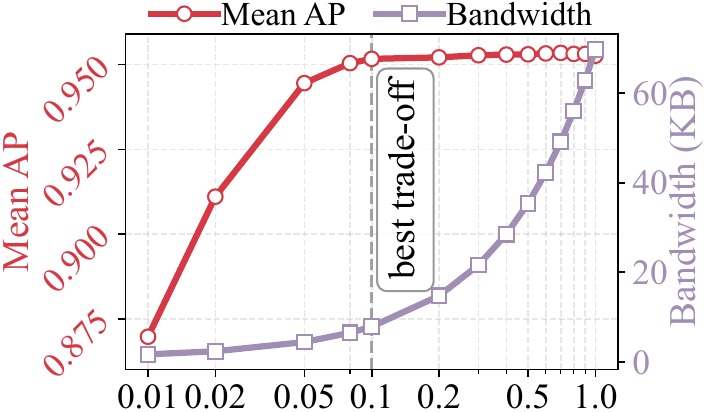}
        \caption{Trade-off w.r.t retention ratio $\alpha$}
        \label{fig:deeper_analysis_sub2}
    \end{subfigure}
    \hfill
    \begin{subfigure}[b]{0.33\textwidth}
        \centering
        \includegraphics[width=\textwidth]{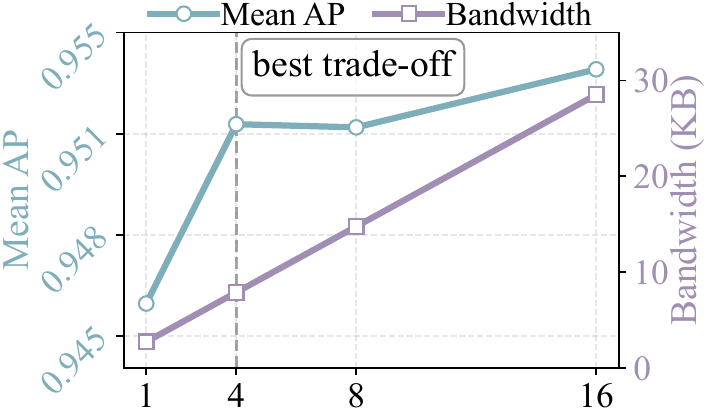}
        \caption{Trade-off w.r.t quantization bit-width $b$}
        \label{fig:deeper_analysis_sub3}
    \end{subfigure}
    \caption{Runtime and trade-off analysis for InfoCom.}
    \label{fig:deeper_analysis}
\end{figure*}

\subsection{Main Results}

\subsubsection{Comparison of Communication-Efficient Methods.} As summarized in Tab. \ref{table:main_result_1}, experimental results reveal three superiorities inherent to InfoCom.
i) \textit{Exceptional communication efficiency}: InfoCom requires only kilobyte-level communication volume, comparable to Late Collaboration but significantly lower than other feature-based solutions. Specifically, its bandwidth consumption is over 400 times lower than Where2comm, only 1\% of that of ERMVP, and over 4000 times lower than Standard Collaboration.
ii) \textit{Superior perception performance}: InfoCom maintains perception performance on par with the bandwidth-intensive Standard Collaboration despite minimal communication overhead, while significantly outperforms Where2comm. Moreover, ERMVP exhibits the smallest performance gap relative to InfoCom.
iii) \textit{Optimal communication-performance trade-off}: InfoCom demonstrates state-of-the-art efficiency in performance gain per unit bandwidth. For example, on the OPV2V dataset, InfoCom achieves $1.8 \times 10^{-2}$ average performance gain per kilobyte, substantially exceeding Where2comm ($3.2 \times10^{-5}$) and ERMVP ($1.7 \times 10^{-4}$).



\subsubsection{Validation of Different Collaborative Models.} To evaluate compatibility, we integrated communication-efficient methods into the base AttFuse and MKD-Cooper models, with results presented in Tab. \ref{table:main_result_2}. The findings reaffirm core conclusions in Tab. \ref{table:main_result_1} while also revealing a notable performance enhancement. For example, the InfoCom-integrated MKD-Cooper variant surpassed the original model by 1.27\% in mean AP. Although AttFuse and MKD-Cooper generate suboptimal single-scale intermediate features compared to CoAlign's multi-scale representations, InfoCom's information purification mechanism enhances feature quality by simultaneously suppressing noise interference and extracting task-critical information. This approach effectively compensates for the feature constraints inherent in weaker collaborative perception backbones.

\subsubsection{Qualitative Analysis.} The visualization results in Fig. \ref{figure:qualitative_analysis} further demonstrate that InfoCom achieves better 3D object detection. Specifically, it outperforms existing methods in prediction accuracy, localization precision, and false positive suppression. These intuitive results are highly consistent with the quantitative analysis presented earlier.

\begin{figure*}[ht]
    \centering
    \includegraphics[width=0.99\linewidth]{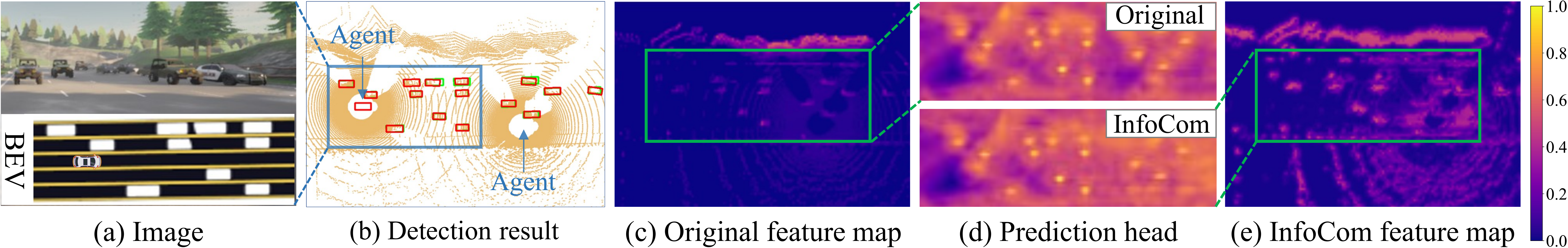}
    \caption{Deeper visualization analysis for InfoCom.}
    \label{fig:deeper_analysis_feature}
\end{figure*}

\begin{table}[t]
    \setlength{\tabcolsep}{2.5mm}
    \centering
    \begin{tabular}{c|c|c|c}
    \toprule
        ID & Component & Variant & Mean AP \\ 
    \midrule
        1 & / & InfoCom (Full) & 0.9518 \\ 
    \midrule
        2 & IAE & Simple Encoder & 0.9320 \\
    \midrule
        3 & \multirow{2}{*}{SMG} & Simple Generator & 0.9379 \\ 
        4 & ~ & w/o STE & 0.8845 \\ 
    \midrule
        5 & \multirow{2}{*}{MSD} & w/o Mask & 0.8839  \\ 
        6 & ~ & w/o Multi-Scale Rec. & 0.9439 \\

    \bottomrule
    \end{tabular}
    \caption{Effects of the different components on InfoCom.}
    \label{tab:ablation_study}
\end{table}

\subsection{Deeper Analysis}

\subsubsection{Runtime Analysis.}
Using identical hardware configurations with a 3090 GPU, we evaluated additional execution time across varying agent densities on the OPV2V. The resultant time is reported in Fig. \ref{fig:deeper_analysis_sub1}, derived from 20 experimental trials. The results show that Where2comm achieves the shortest computation time, approximately half of InfoCom, whereas ERMVP incurs the most substantial computational overhead. Note that total system latency consists of both data transmission time and computation time. Where2comm's MB-level data transmission demands substantially offset its computational advantages. Conversely, InfoCom's KB-level communication overhead demonstrates significant potential for overall latency optimization.

\subsubsection{Validation of Spatial Cue Sparsity.}
\label{sec:sub_saptial_cus}

By progressively adjusting the retention ratio of sparse mask $M$ in Fig. \ref{fig:deeper_analysis_sub2}, we empirically demonstrate that transmitting more than 10\% of spatial cues yields only marginal performance gains ($<0.2 \%$ AP on average). This results from the inherent sparsity of point cloud data. Accordingly, controlling the sparsity of bandwidth-dominant $M$ is essential for optimizing InfoCom's communication efficiency.

\subsubsection{Feasibility of Low-Precision Representation.}
Fig. \ref{fig:deeper_analysis_sub3} reveals redundancy in high-precision floating-point representations of spatial cues. Experiments demonstrate that quantizing $M$ to $4$-bit precision reduces communication overhead by a factor of 8 while slightly affecting perception performance (AP fluctuation $<0.18\%$). Critically, InfoCom's uniform quantization strategy incurs negligible computational costs and additional burden.

\subsubsection{Feature-Level Comparative Visualization.}

Fig. \ref{fig:deeper_analysis_feature} visualizes feature disparities in a two-agent scenario, depicting: (a) the perspective of the right agent from Fig. \ref{fig:deeper_analysis_feature}b, (b) collaborative perception results, and (c-e) feature comparisons between standard collaboration and InfoCom, with yellow regions indicating higher activation values. InfoCom's task-driven information purification mechanism fundamentally redefines information transmission by conveying KB-scale critical perception data rather than spatially structured key features. This approach eliminates explicit feature-level alignment constraints, resulting in significant feature map disparities evident in Figs. \ref{fig:deeper_analysis_feature}c and \ref{fig:deeper_analysis_feature}e, yet maintains uncompromised perception performance. As shown in Fig. \ref{fig:deeper_analysis_feature}d, both prediction heads exhibit consistent response intensities in target regions while equally suppressing background.


\subsection{Ablation Study}

Tab. \ref{tab:ablation_study} summarizes the ablation study on InfoCom's three key components: IAE, SMG, and MSD. The first row, representing the complete framework, serves as the baseline. Rows 2, 3, and 5 evaluate simplified implementations of individual components. These results demonstrate a positive contribution for each key element. Specifically, rows 2 and 3 indicate that the simplified variants of IAE and SMG underperform relative to the complete modules. Row 4 reveals that the non-differentiability introduced by quantization and filtering operations during joint compression post-processing necessitates an appropriate gradient estimation technique. Row 5 indicates that spatial cues effectively mitigate information loss under aggressive compression. Finally, row 6 shows that replacing Multi-Scale Decoding with single-masked reconstruction in the final stage leads to marginal performance degradation.

\section{Conclusion}
We propose InfoCom, a novel communication-efficient framework for collaborative perception. Compared to existing methods, it delivers three distinct advantages: reducing communication volume from megabytes to kilobytes by condensing perception-critical information rather than manipulating redundant spatial features; providing theoretical analysis based on information principles to address the empirical limitations of heuristic designs; and enabling plug-and-play functionality via a standardized modular architecture. These advantages originate from three complementary innovations: Information-Aware Encoding, Sparse Mask Generation, and Multi-Scale Decoding. Comprehensive evaluations on OPV2V, V2XSet, and DAIR-V2X datasets consistently demonstrate superior communication efficiency and perception performance.

\section*{Ethical Statement}
We do not foresee ethical concerns posed by our method, but concede that both ethical and unethical applications of autonomous driving techniques may benefit from the improvements induced by our work. Care must be taken, in general, to ensure positive ethical and societal consequences of autonomous driving.

\section*{Acknowledgements}
This work was supported in part by the National Natural Science Foundation of China under Grant 62172342, Grant 62372387; Key R\&D Program of Guangxi Zhuang Autonomous Region, China (Grant No. AB22080038, AB22080039); The Open Fund of the Engineering Research Center of Sustainable Urban Intelligent Transportation, Ministry of Education, China (Project No. KCX2024-KF07).

\bibliography{aaai2026}

\clearpage

\section{Appendix of InfoCom}

\setcounter{secnumdepth}{1}
\renewcommand{\thesection}{\Alph{section}}

\section{Contributions and Future Directions}
\label{sec:appendix_contributions}

\subsection{Highlights of Our Contribution}

In summary, our contributions are three-fold.
\begin{itemize}
\item We propose InfoCom, the first theoretically guaranteed communication-efficient collaborative perception framework. It resolves the fundamental tension between extreme KB-scale compression and perception fidelity through a novel information purification paradigm.

\item InfoCom achieves unprecedented bandwidth efficiency (440×/90× reduction vs. state-of-the-art) while maintaining near-lossless perception accuracy through three co-designed innovations: Information-Aware Encoding, Sparse Mask Generation, and Multi-Scale Decoding.

\item InfoCom demonstrates practical scalability via plug-and-play modularity, which seamlessly integrates with existing collaborative models under real-world network constraints. This capability is validated across OPV2V, V2XSet, and DAIR-V2X datasets.
\end{itemize}

\begin{figure*}[t]
    \centering
    \includegraphics[width=0.85\linewidth]{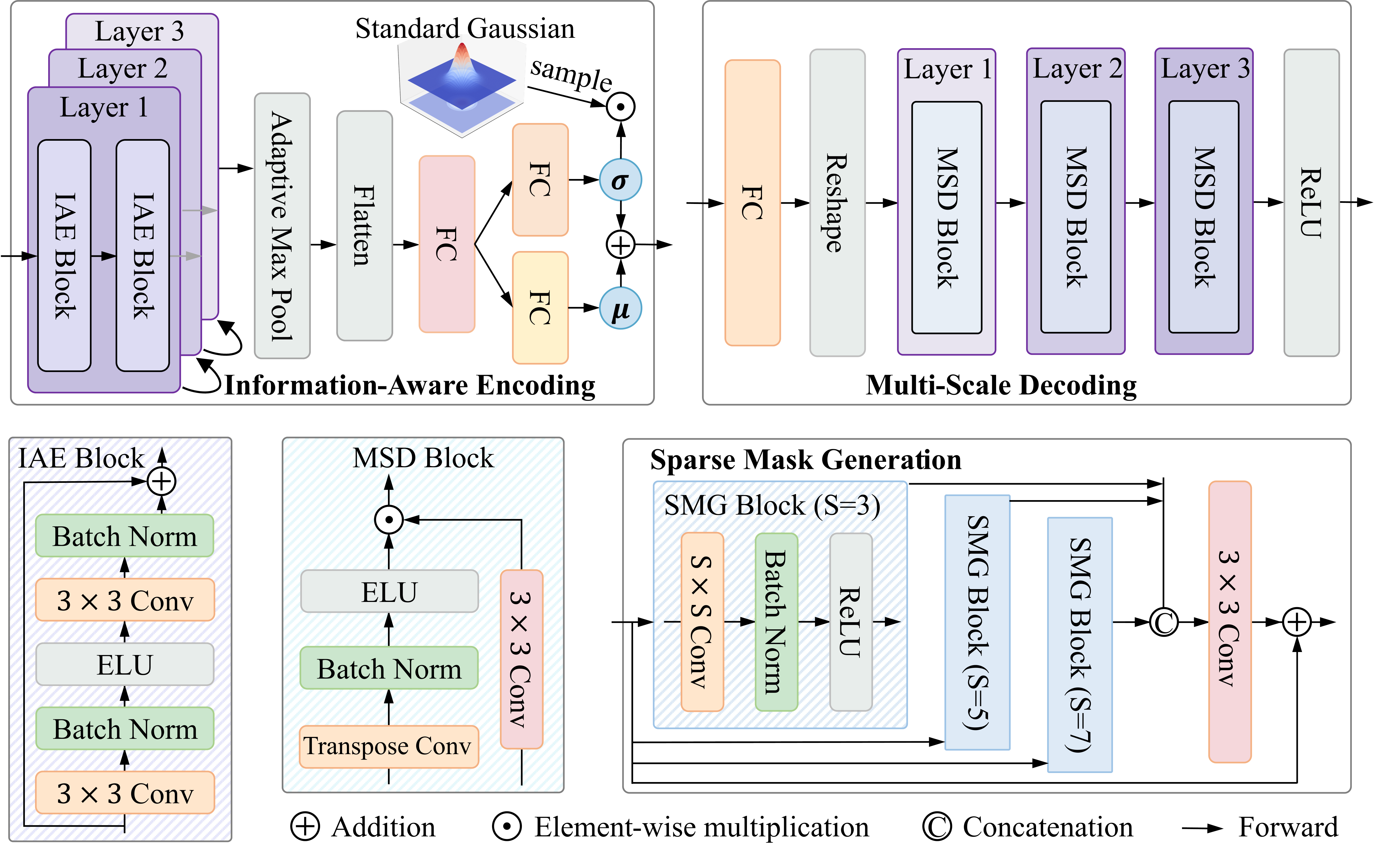}
    \caption{Illustration of the proposed Information-Aware Encoding, Sparse Mask Generation, and Multi-Scale Decoding. }
    \label{fig:networks}
\end{figure*}

\subsection{Limitation and Future Work}
While InfoCom successfully reduces the bandwidth requirements for communication-efficient collaborative perception from megabytes (MB) to kilobytes (KB), the current framework still presents opportunities for further optimization. Specifically, InfoCom's existing message unit comprises a minimal sufficient information-aware feature and a sparse mask that provides contextual spatial cues. Future research will expand this structure into a triple format by incorporating temporal cues. This addition will facilitate dynamic assessments of transmission necessity for collaborative agents in future frames.

\section{Network Architectures}
\label{sec:appendix_architectures}

Fig. \ref{fig:networks} details the network architectures of the three core components in the InfoCom framework. The overall system architecture is illustrated in Fig. 2 (main text). 

\subsection{Information-Aware Encoding (IAE)}
The IAE module comprises three encoding layers, each containing two residual-style basic blocks. Features from the final encoding layer sequentially pass through an adaptive max pooling, a flattening, and a fully connected layer. Two parallel fully connected layers subsequently estimate the mean $\mu$ and standard deviation $\sigma$ of the Gaussian distribution representing latent features. Deployed on each transmitter agent, this module extracts information-aware features $E$ from intermediate features $Z$, minimizing transmission bandwidth while preserving information critical for downstream tasks. Finally, differentiable sampling is enabled via the reparameterization trick, which defines the latent representation as $E = \mu \oplus \sigma \odot \epsilon$ with $\epsilon \sim \mathcal{N}(0, I)$.

\subsection{Sparse Mask Generation (SMG)}
The SMG module at the sender end compensates for critical spatial cues lost during IAE's aggressive compression by providing auxiliary spatial information. SMG primarily consists of multiple basic blocks based on convolutional kernels of varying sizes. The module takes the $Z$ as input and outputs a spatial mask that captures spatial cues. To minimize transmission bandwidth, the spatial mask is post-processed by filtering and quantization, resulting in the sparse mask $M^q$. In our implementation, three SMG blocks with different convolutional kernel sizes extract spatial information at varying granularities. The extracted features are then concatenated along the channel dimension. Subsequently, a convolution processes the concatenated features to adjust the channel to the target number. Residual connections are optional in this process.

\subsection{Multi-Scale Decoding (MSD)}
The MSD module operates at the receiver end, reconstructing actionable Bird's-Eye View (BEV) features from transmitted 1D features $E$ and sparse mask $M^q$. Its core comprises a series of mask-guided MSD blocks that progressively reconstruct low-resolution BEV features into higher-resolution representations through $M^q$-modulated feature refinement. The processing flow involves: i) a fully connected layer expanding $E$'s dimensionality and reshaping it into 2D initialization features; ii) decoding layers with three fundamental MSD blocks incrementally reconstructing features to the target $C\times H \times W$ BEV space; and iii) a ReLU activation matching the sparsity characteristics of original intermediate feature $Z$.

In summary, InfoCom's network architecture prioritizes lightweight design and operational efficiency. Experimental results demonstrate an additional inference latency of approximately 0.02 seconds, which is significantly lower than the 0.16 seconds required by the state-of-the-art ERMVP method. Through synergistic interaction of its three core components, InfoCom condenses collaborative messages with minimally sufficient information for perception tasks. Consequently, this approach reduces communication overhead to kilobytes while maintaining competitive perception performance, significantly improving communication efficiency in collaborative perception systems.



\section{Experiment Details}
\label{sec:appendix_experiment_details}

\subsection{Hardware and Software Configurations}
All experiments in this paper were conducted on a single device with the following hardware and software configurations.

\begin{itemize}
    \item Ubuntu 20.04.6 LTS OS, Intel(R) Xeon(R) Silver 4314 CPU @ 2.40GHz, 64 GB memory, and RTX 3090 GPU.
    \item CUDA 12.2, Python 3.7.16, Pytorch 1.10.1, spconv 2.3.6, and OpenCOOD 0.1.0.
\end{itemize}

\subsection{Implementation Details}
We adopt PointPillars as the detector that is consistent with prior work. \cite{Lu2022RobustC3}. Initially, we train the collaborative perception model for 30 epochs with a learning rate of 0.002 under the default configuration. Subsequently, three efficient communication methods are integrated into the optimal model and fine-tuned for an additional 30 epochs, with reporting peak performance. For intermediate feature resolution, CoAlign uses $64\times200\times704$, while AttFuse and MKD-Cooper operate at $256\times100\times352$. In InfoCom, the feature $E$ for transmission has 256 channels. We set the Lagrange multiplier $\beta$ to $0.01$ and the default quantization bandwidth $b$ to $4$ bits.

\subsection{Communication Volume Calculation}

Our communication volume calculation builds upon existing efforts but uses intuitive units (KB or MB) rather than logarithmic scaling to highlight bandwidth consumption differences. Total communication volume aggregates the sizes of all collaborative message units for each agent, including primary messages and auxiliary information. For InfoCom, the communication volume is calculated as: 

\begin{equation}
    \bigl((D \times 32) + (\alpha \times H \times W \times b)\bigr) / 8.
\end{equation}

Here, the first term is the communication volume of the information-aware feature $E$, and the second term corresponds to the size of the sparse mask $M^q$. $D$ denotes the dimension of $E$, $H \times W$ is the resolution of $M^q$, $32$ indicates the default float32 data type, $\alpha$ is the retention rate in the filtering step of Eq. 6, and $b$ represents the quantization bit-width in Eq. 7. Additionally, division by $8$ converts the unit from bits to bytes. Note that in InfoCom, the primary message $E$ does not undergo quantization similar to $M^q$, as additional quantization would yield only marginal bandwidth benefits while compromising perception performance.

\subsection{Setup of Ablation Study}
Detailed descriptions of ablation variants used in Tab. 3 are as follows:

\begin{enumerate}

\item \textbf{InfoCom (Full)} is the complete InfoCom without any component removal, serving as the baseline.
\item \textbf{Simple Encoder} is a variant that replaces the Information-Aware Encoding with a weaker Naive Encoder (a simple block comprising $3 \times 3$ convolution and activation functions), while retaining other designs.
\item \textbf{Simple Generator} retains only the first SMG block, which eliminates multi-granularity spatial modelling.
\item \textbf{w/o STE} ablates the Straight-Through Estimator (STE), which disrupts gradient backpropagation through non-differentiable mask filtering and quantization operations.
\item \textbf{w/o Mask} removes mask-guided feature reconstruction in MSD blocks, validating extreme compression effects on perception performance without spatial guidance.
\item \textbf{w/o Multi-Scale Rec.} restricts mask-guided reconstruction to the final layer, which is a stronger variant than 5.

\end{enumerate}

\section{More Experiments}
\label{sec:appendix_more_experiments}

\begin{table}[th]
    \setlength{\tabcolsep}{1.1mm}
    \centering
    \begin{tabular}{c|rrrr}
    \toprule
        Comm. Method & 2 & 3 & 4 & 5 \\ 
    \midrule
        Standard Colla. (MB) & 34.375 & 68.750 & 103.125 & 137.500  \\ 
        Where2comm (MB) & 3.439 & 6.878 & 10.318 & 13.757 \\ 
        ERMVP (MB) & 0.741 & 1.482 & 2.224 & 2.965  \\ 
    \midrule
        InfoCom (KB) & 7.875 & 15.750 & 23.625 & 31.500 \\ 
    \bottomrule

    \end{tabular}
    \caption{Communication volume vs. agent density.}
    \label{tab:Communication_volume}
\end{table}

\begin{figure}[!th]
    \centering
    \includegraphics[width=0.9\linewidth]{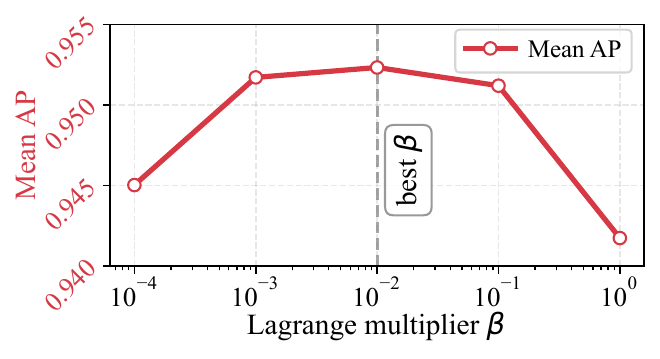}
    \caption{Sensitivity analysis of Lagrange multiplier $\beta$.}
    \label{fig:Parameter_sensitivity}
\end{figure}

\subsection{Communication Volume under Varying Agent Number}

As Tab \ref{tab:Communication_volume} shows, InfoCom consistently achieves the lowest communication overhead in experiments with 2 to 5 agents, outperforming standard collaborative perception and two feature-based communication-efficient methods. This advantage persists across sparse and dense scenarios, enabling stable operation under real-world 5G-V2X bandwidth constraints (e.g., 0.4 MB/s) \cite{raca2020beyond}. InfoCom exhibits linear communication growth while maintaining kilobyte-scale volumes, demonstrating potential for real-time collaborative perception in traffic-intensive environments.

\subsection{Parameter Sensitivity Analysis}
Fig. \ref{fig:Parameter_sensitivity} presents the sensitivity analysis of hyperparameter $\beta$ in the InfoCom loss function, which balances maximization of perception-relevant information against minimization of redundant information. Optimal performance is achieved at $\beta=10^{-2}$, with unexpectedly low fluctuations in performance across $\beta$ values. $\beta$ solely regulates the information trade-off during kilobyte-scale message generation in the IAE module, while the SMG module independently provides critical spatial cues. This decoupled extraction of spatial and semantic information may explain $\beta$'s low sensitivity.

\end{document}